\documentclass[letterpaper, 10 pt, journal, twoside]{ieeetran}
\IEEEoverridecommandlockouts
\usepackage{graphicx}
\usepackage{amsmath, amssymb, amsfonts}

\usepackage{amsthm}
\usepackage{bm}
\usepackage{algorithm}
\usepackage[noend]{algpseudocode}
\usepackage{textcomp}
\usepackage{comment}
\usepackage[caption=false]{subfig}
\usepackage[table]{xcolor}
\usepackage{cite}
\usepackage{outlines}
\usepackage{tablefootnote}

\usepackage{color}

\newcommand{\rev}[1]{#1}
\newcommand{\finrev}[1]{#1}

\usepackage{braket}
\usepackage{mathtools}
\DeclarePairedDelimiter{\abs}{\lvert}{\rvert}

\theoremstyle{definition}

\newtheorem{thm}{Theorem}

\newtheorem{prb}{Problem}

\algrenewcommand\algorithmicrequire{\textbf{Input:}}
\algrenewcommand\algorithmicensure{\textbf{Output:}}

\usepackage{xspace}
\newcommand{\tlnet}{\textsc{TLNet}\xspace}

\usepackage[hidelinks]{hyperref}
\hypersetup{
    colorlinks=false,
    citecolor=blue,
    linkcolor=blue,
    filecolor=blue,
    urlcolor=blue,
    pdfpagemode=FullScreen,
}
\usepackage{
    cleveref
} 

\usepackage[inline]{enumitem}

\begin{document}

    \title{Neuro-Symbolic Generation of Explanations for Robot Policies with
    Weighted Signal Temporal Logic}



    \author{\finrev{Mikihisa Yuasa$^{1}$,  Ramavarapu S. Sreenivas$^{1}$, Huy T. Tran$^{1}$}%
\thanks{\finrev{Manuscript received: May, 30, 2025; Revised October, 14, 2025; Accepted January, 28, 2026.
This paper was recommended for publication by Editor Editor A. Bera upon evaluation of the Associate Editor and Reviewers’ comments.
This work was supported in part by the Office of Naval Research (ONR) under Grant N00014-20-12249 and JASSO Study Abroad Fellowship Program (Graduate Degree Program).}}
\thanks{\finrev{$^{1}$The Grainger College of Engineering, University of Illinois Urbana-Champaign, Urbana, IL 61801 USA.
        {\tt\small \{myuasa2,rsree,huytran1\}@illinois.edu}}}%
\thanks{\finrev{Digital Object Identifier (DOI): see top of this page.}}
}

    \maketitle

    \begin{abstract}
        Learning-based policies have demonstrated success in many robotic applications, but often lack explainability. 
        We propose a neuro-symbolic explanation framework that generates a weighted signal temporal logic (wSTL) specification which describes a robot policy in a human-interpretable form. 
        Existing methods typically produce explanations that are verbose and inconsistent, which hinders explainability, and are loose, which limits meaningful insights. 
        We address these issues by introducing a simplification process consisting of predicate filtering, regularization, and iterative pruning.
        We also introduce three explainability metrics---conciseness, consistency, and strictness---to assess explanation quality beyond conventional classification accuracy. 
        Our method---\textsc{TLNet}---is validated in three simulated robotic environments, where it outperforms baselines in generating concise, consistent, and strict wSTL explanations without sacrificing accuracy. 
        This work bridges policy learning and explainability through formal methods, contributing to more transparent decision-making in robotics.
    \end{abstract}

    \section{Introduction}
    \label{sec:intro}

    \IEEEPARstart{R}{ecent}
    innovations in learning-based methods, particularly those using neural
    networks, have improved high-level decision-making and
    low-level control in robotics. However, neural network policies are
    inherently black-box in nature, which poses significant concerns for safety-critical
    robotic applications, such as autonomous vehicles, where explainability is paramount.
    Various approaches have been explored to enhance the explainability
    of learned policies \cite{milani_explainable_2024},
    including post-hoc techniques
    \cite{iucci_explainable_2021} and methods to learn
    policies that are inherently explainable
    \cite{kenny_interpretable_2022}.

    In this paper, we focus on post-hoc analysis to enable explanation of any given
    robot policy. 
    While existing approaches, like saliency maps \cite{mott_interpretable_2019}, feature attribution \cite{puri_explain_2019}, and \rev{non-temporal concept/rule extraction \cite{ragodos_protox_2022},} provide valuable information, they often lack language-based explanations, which limits human interpretability.
    Autoregressive language models have shown promising results for providing
    language-based explanations \cite{trigg_natural_2024},
    but they do not support formal verification of the output explanation, which
    limits their applicability in safety-critical domains \cite{buzhinsky_formalization_2019}.

    Temporal logic offers an alternative approach to specifying robotic tasks and system constraints in an interpretable yet mathematically precise manner that supports verification \cite{cosler_nl2spec_2023,bartocci_survey_2022,bhuyan_neuro-symbolic_2024}.
    Recent efforts have also used temporal logic to infer explanations of given
    trajectory data. For example,
    \cite{gaglione_learning_2021,bombara_offline_2021}
    use a decision tree \rev{(DT)} classifier to infer the linear temporal logic (LTL) or signal temporal logic (STL) specification that
    best explains the difference between two classes of trajectories.
    \cite{yuasa_generating_2024} uses a greedy search over candidate LTL explanations driven by policy similarity metrics. However, these methods
    are limited in their ability to infer specifications where certain predicates
    are more critical than others. \textit{Weighted STL} (wSTL) \cite{mehdipour_specifying_2021} extends STL by
    introducing importance weights over predicates, which can highlight the
    relative priority of various task components. The expressiveness of wSTL thus
    makes it particularly suitable for describing robotic behaviors in settings with
    competing or hierarchical objectives.

    \begin{figure}[tb]
        \centering
        \includegraphics[width=0.65\linewidth]{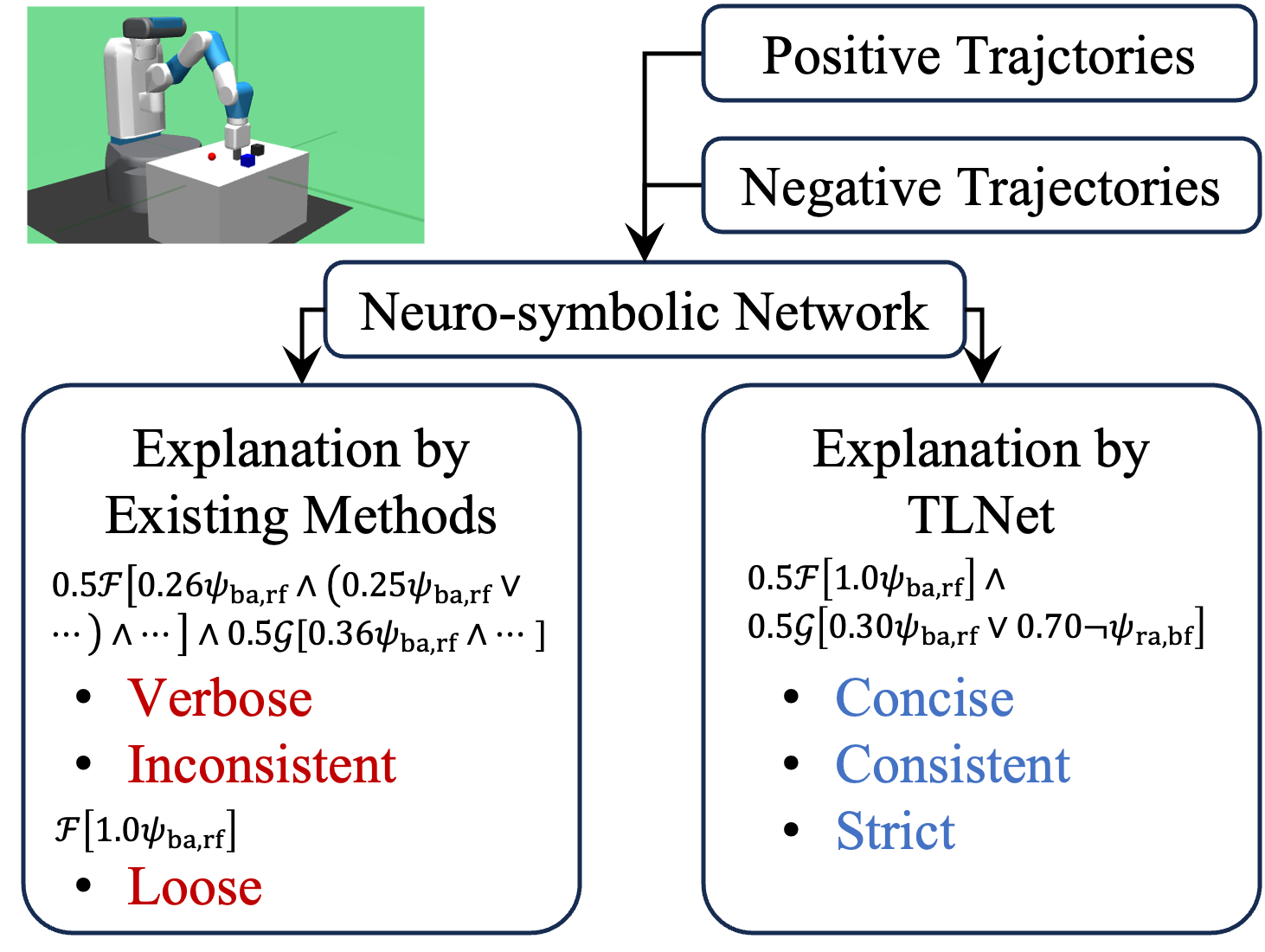}
        \caption{
            Existing neuro-symbolic methods often produce explanations that are verbose, inconsistent, and loose. Our method, \tlnet, addresses this by generating concise, consistent, and strict wSTL specifications.
        }
        \label{fig:comparison}
    \end{figure}

    The presence of real-valued weights in wSTL has enabled the use of neural
    networks for inferring wSTL specifications that explain given
    trajectory data. These methods have shown
    high classification accuracy for various applications, including anomaly detection
     \cite{yan_stone_2021}, fault diagnosis
    \cite{tian_neural-symbolic_2024}, and weather forecasting \cite{baharisangari_weighted_2022}.
    However, existing work only focuses on classification accuracy, which does
    not fully capture the actual explainability of learned specifications. For example,
    as shown in \Cref{fig:comparison}, these methods often produce specifications
    that are unnecessarily verbose (i.e., they include many unnecessary
    predicates), inconsistent across seeds, and/or loose, such that they are
    satisfied but do not provide useful insights (e.g., ``catch-all'' specifications
    that include many conjunctions or overly simplified specifications). Some works address verbosity by incorporating
    heuristic simplifications, like only keeping the top-$k$ 
    predicates \cite{yan_stone_2021,yan_neuro-symbolic_2022} or using a greedy
    process to remove weights \cite{li_learning_2023}. However, no existing methods effectively balance classification
    accuracy with verbosity, inconsistency, and looseness.

    In this work, we propose \tlnet, a neural network architecture for inferring
    wSTL explanations of robot policies that are concise (i.e., include as few
    predicates as needed), consistent (i.e., produce similar explanations across
    training seeds), and strict (i.e., contain as few disjunctions as needed).
    \tlnet achieves this through a multi-step simplification process consisting of predicate filtering, a regularized loss function, and iterative weight pruning.
    We provide theoretical justification for our network's weight distributivity and introduce explainability metrics---conciseness, consistency, and strictness---to complement standard classification accuracy.
    We demonstrate our method in three simulated robotics environments, showing it outperforms baselines \rev{in our proposed metrics while matching or improving classification accuracy.}

    \section{Preliminaries}
    \label{sec:bg}

    Let a trajectory (or a signal) of $n$-dimensional states be a discrete-time
    series $\tau = s_{0}, s_{1}, \dots, s_{H}$, where $s_{t}\in S$ is a state at
    timestep $t \in \mathbb{Z}_{\geq0}$, $S \subseteq \mathbb{R}^{n}$ is the state
    space, and $H$ is the time horizon of $\tau$. We also define an interval $I$
    as
    $I:=[a,b] = \{k \in \mathbb{Z}_{\geq0}| a\leq k \leq b; a,b\in\mathbb{Z}_{\geq0}
    \}$, $\abs{I}$ as the cardinality of $I$, and $t+I$ as the interval
    $[t+a, t+b]$.


    \subsection{Weighted Signal Temporal Logic}

We consider wSTL syntax \cite{mehdipour_specifying_2021} defined recursively as
    $\phi := \top \mid \psi \mid \neg\phi \mid 
    \bigwedge_{i=1}^N{}^{w} \phi_i \mid
    \bigvee_{i=1}^N{}^{w} \phi_i \mid \mathcal{G}_I^{w} \phi \mid \mathcal{F}_I^{w} \phi$,
where $\top$ is the logical \textit{True} value, $\psi := f(s) \geq c$ is an atomic predicate given a function $f:S \rightarrow \mathbb{R}$ with constant $c \in \mathbb{R}$, and $\neg$, $\wedge$, and $\vee$ are Boolean \textit{negation}, \textit{conjunction}, and \textit{disjunction} operators, respectively.
\rev{$\mathcal{G}_I$ and $\mathcal{F}_I$ are \textit{globally} and \textit{eventually} operators, respectively.}
For conjunctive and disjunctive operators, we define the weights as $w:=[w_i]_{i=1:N} \in \mathbb{R}^N_{>0}$, where $N$ is the number of subformulae in the operator. For temporal operators, we define the weights as $w := [w_k]_{k\in I} \in \mathbb{R}^{\abs{I}}_{>0}$ for interval $I$. 
For notational simplicity, we use the following notation interchangeably: $\bigwedge_{i=1}^N{}^{w} \phi_i = \bigwedge_{i=1}^N w_i\phi_i$ and $\bigvee_{i=1}^N{}^{w} \phi_i = \bigvee_{i=1}^N w_i \phi_i$.

    \subsection{Normalized wSTL}

    Our neural network architecture proposed in \Cref{subsec:nn-architecture}
    requires the weights within each operator to be normalized. We discuss the reasons
    for this requirement in \Cref{subsec:theory}. Given this requirement, we define
    normalized wSTL syntax recursively as $\phi := \top \mid \psi \mid \neg\phi \mid
    \bigwedge_{i=1}^{N}{}^{w}\phi_{i}\mid \bigvee_{i=1}^{N}{}^{w}\phi_{i}\mid \mathcal{G}
    _{I}^{w}\phi \mid \mathcal{F}_{I}^{w}\phi,$ where all terminology carries
    over from wSTL, except we further require
    $\sum_{i=1}^{N}w_{i}= \sum_{k\in I}w_{k}= 1$, where $w_{i}, w_{k}\in [0,1 ]$. 
    We allow the weights to be zero because the zero weights do not influence the
    overall robustness \cite{yan_stone_2021}.

    To avoid large input values to our neural network, we also define normalized
    quantitative satisfaction semantics, or \emph{robustness}, by extending \cite{mehdipour_specifying_2021}
    as follows,

    \begin{align}
        & r(\psi, \tau, t)                                         & := & \begin{cases}\frac{f(s_{t}) - c }{\sup_{s\in S}f(s) - c} & \text{if $f(s_{t}) - c \geq 0$,}\\ \frac{f(s_{t}) - c }{\inf_{s \in S}f(s) - c} & \text{otherwise}\end{cases},\label{eqn:bounds} \\
         & r(\neg\phi, \tau, t)                                     & := & -r(\phi, \tau, t), \\
         & r\left(\bigwedge_{i=1}^{N}{}^{w}\phi_{i}, \tau, t\right) & := & \otimes^{\wedge}(w, [r(\phi_{i},\tau, t)]_{i = 1:N}),       \\
         & r\left(\bigvee_{i=1}^{N}{}^{w}\phi_{i}, \tau, t\right)   & := & \oplus^{\vee}(w, [r(\phi_{i}, \tau, t)]_{i = 1:N}), 
    \end{align}
    \begin{align}
         & r\left(\mathcal{G}^{w}_{I}\phi, \tau, t \right)          & := & \otimes^{\mathcal{G}}_{I}(w, [r(\phi,\tau, t')]_{t' \in t + I }),\\
         & r\left(\mathcal{F}^{w}_{I}\phi, \tau, t \right)          & := & \oplus^{\mathcal{F}}_{I}(w, [r(\phi, \tau, t')]_{t' \in t + I }),
    \end{align}
    where the aggregation functions
    $\otimes^{\wedge}:\mathbb{R}^{N}_{\geq0}\times\mathbb{R}^{N}\rightarrow\mathbb{R}
    , \oplus^{\vee}:\mathbb{R}^{N}_{\geq0}\times\mathbb{R}^{N}\rightarrow\mathbb{R}
    , \otimes^{\mathcal{G}}_{I}:\mathbb{R}^{\abs{I}}_{\geq0}\times\mathbb{R}^{\abs{I}}
    \rightarrow\mathbb{R}$, and $\oplus^{\mathcal{F}}_{I}:\mathbb{R}^{\abs{I}}_{\geq0}
    \times\mathbb{R}^{\abs{I}}\rightarrow\mathbb{R}$ correspond to the operators
    $\wedge, \vee, \mathcal{G}$, and $\mathcal{F}$, respectively. \rev{In practice, the infimum and supremum bounds in \Cref{eqn:bounds} can be inferred from physical properties of the environment (we assume known map sizes in our results) or sampled trajectories.}
    For notational simplicity, we interchangeably use $r(\phi,\tau,t)$ and
    $r_{\tau t}^{\phi}$.

    Building from \cite{yan_stone_2021}, we define the aggregation functions with
    a scaling constant $\sigma > 0$ as follows,
    \begin{align}
        \otimes^{\wedge}(w, [r_{\tau t}^{\phi_i}]_{i = 1:N})           & := \frac{\sum_{i=1}^{N}w_{i}r_{\tau t}^{\phi_i}\exp(\frac{-r_{\tau t}^{\phi_i}}{\sigma})}{\sum_{i=1}^{N}w_{i}\exp(\frac{-r_{\tau t}^{\phi_i}}{\sigma})},  \label{eqn:agg_and}     \\
        \oplus^{\vee}(w, [r_{\tau t}^{\phi_i}]_{i = 1:N})              & := \frac{\sum_{i=1}^{N}w_{i}r_{\tau t}^{\phi_i}\exp(\frac{r_{\tau t}^{\phi_i}}{\sigma})}{\sum_{i=1}^{N}w_{i}\exp(\frac{r_{\tau t}^{\phi_i}}{\sigma})},   \label{eqn:agg_or}      \\
        \otimes^{\mathcal{G}}_{I}(w, [r_{\tau t'}^{\phi}]_{t' \in t + I }) & := \frac{\sum_{t' \in t + I}w_{t'}r_{\tau t'}^{\phi}\exp(\frac{-r_{\tau t'}^\phi}{\sigma})}{\sum_{t' \in t + I}w_{t'}\exp(\frac{-r_{\tau t'}^\phi}{\sigma})}, \label{eqn:agg_g} \\
        \oplus^{\mathcal{F}}_{I}(w, [r_{\tau t'}^{\phi}]_{t' \in t + I })  & := \frac{\sum_{t' \in t + I}w_{t'}r_{\tau t'}^{\phi}\exp(\frac{r_{\tau t'}^\phi}{\sigma})}{\sum_{t' \in t + I}w_{t'}\exp(\frac{r_{\tau t'}^\phi}{\sigma})}.   \label{eqn:agg_f} 
    \end{align}
    \rev{We empirically found that decreasing $\sigma$ better approximates the max/min aggregations, but can increase training instability due to larger gradients, while increasing $\sigma$ too much can lead to nonsensical explanations.}

    \begin{figure*}[tb]
        \medskip
        \centering
        \includegraphics[width=0.80\linewidth]{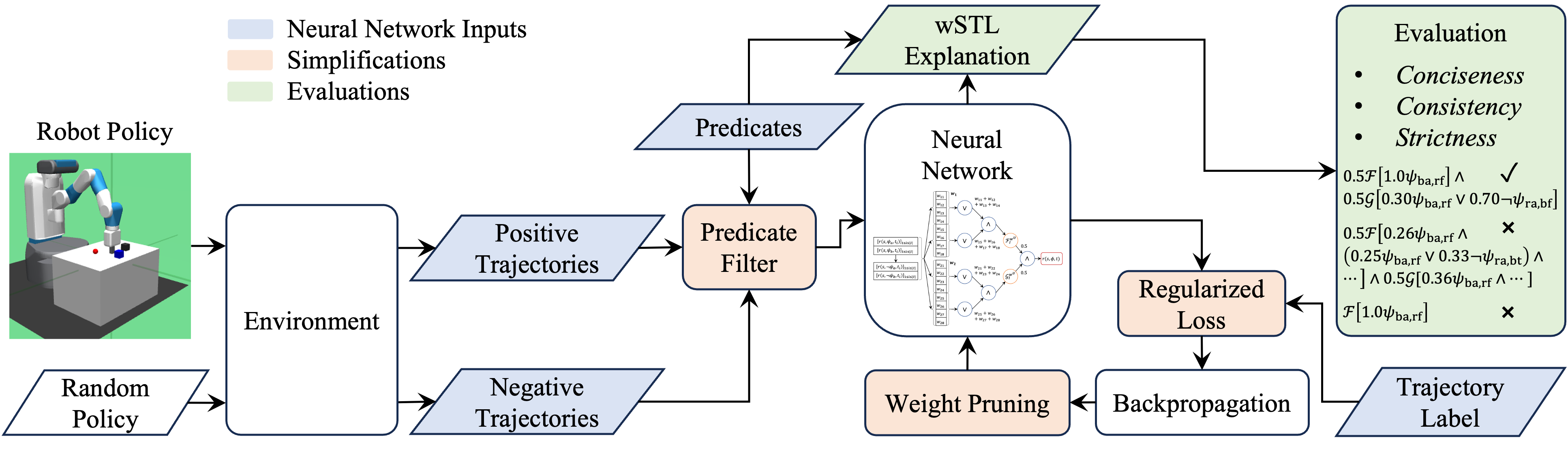}
        \caption{We propose a neuro-symbolic approach for generating an
        explanation of a given robot policy using wSTL. Our network architecture,
        \tlnet, includes a simplification process that ensures resulting explanations
        balance classification accuracy with conciseness, consistency, and strictness.}
        \label{fig:overview}
    \end{figure*}

    \subsection{Problem Statement}

    We consider the following binary classification task. Let $\mathcal{T}= \Set{(\tau_i, l_i)}
    _{i=1:N_\mathrm{data}}$ be a labeled dataset, where
    $\tau_{i}\in\mathbb{R}^{n \times H}$ is the $i^{th}$ trajectory with label
    $l_{i}\in \{1, -1\}$ and $N_{\mathrm{data}}\in \mathbb{Z}_{>0}$ is the
    number of data points in the dataset.
    \begin{prb}
        Given a dataset $\mathcal{T}$, find a wSTL specification $\phi$ that
        balances classification accuracy with conciseness, consistency, and
        strictness.
    \end{prb}

    \section{Methodology}
    \label{sec:method}
    We first introduce our neural network architecture, \tlnet, which allows us
    to represent wSTL structures in a fully differentiable form. We then introduce
    our simplification process, which enhances the explainability of inferred
    explanations, and our trajectory sampling process. Finally, we propose novel
    evaluation metrics for explainability---\emph{conciseness}, \emph{consistency},
    and \emph{strictness}---and provide theoretical justification for our
    network architecture.
    \Cref{fig:overview} and \Cref{alg:expl} summarize our overall approach.

    \subsection{Neural Network Architecture}
    \label{subsec:nn-architecture}

    Let $\Psi =\Set{\psi_i}_{i=1: N_\mathrm{AP}}$ be a predefined set of atomic predicates and $\Psi_{\neg}= \Set{\neg\psi_i}_{i=1: N_\mathrm{AP}}$ be a set of their negations.
    We focus on representing normalized wSTL structures composed of temporal operators containing these predicates in conjunctive normal form (CNF), as follows,
    \begin{align}
        \phi = \notag & 0.5\mathcal{F}_{I}^{\bar{w}}\left[\bigwedge_{i=1}^{N_\mathrm{AP}}w_{i}^{\mathcal{F}}\bigvee_{j=1}^{2N_\mathrm{AP}}w_{ij}^{\mathcal{F}}\psi_{j}\right]                                \\
        \wedge        & 0.5\mathcal{G}_{I}^{\bar{w}}\left[\bigwedge_{i=1}^{N_\mathrm{AP}}w_{i}^{\mathcal{G}}\bigvee_{j=1}^{2N_\mathrm{AP}}w_{ij}^{\mathcal{G}}\psi_{j}\right], \label{eqn:nn_templ_standard}
    \end{align}
    where
    $w_{i}^{\mathcal{F}}, w_{ij}^{\mathcal{F}}, w_{i}^{\mathcal{G}}, w_{ij}^{\mathcal{G}}
    \in [0, 1]$
    are importance weights and $\psi_{j}\in \Psi \cup \Psi_{\neg}$ are predefined atomic predicates.
    We define the weights $\bar{w}$ for temporal operators over interval $I = [0,H]$ to be $\bar{w}= [\bar{w}_{k}= 1/\abs{I}]_{k\in I}$. 
    We define the conjunction weights of the temporal clauses to be 0.5 to
    equally weigh the importance of the task ($\mathcal{F}$) and constraint ($\mathcal{G}$)
    parts of the explanation. We choose this class of structures because they allow
    us to represent a general class of robot policies aiming to complete a task while
    satisfying a set of constraints \cite{li_formal_2019,yuasa_generating_2024}.

    We represent specifications of the form given in \Cref{eqn:nn_templ_standard}
    by designing a neural network $f_{\mathbf{w}}$ parameterized by $\mathbf{w}$,
    where $\mathbf{w}$ defines the importance weights of the currently
    represented specification $\phi_{\mathbf{w}}$. The input to the neural network
    is a sequence of the robustness values for each predicate $\psi \in \Psi \cup
    \Psi_{\neg}$ for each timestep in a trajectory $\tau$. The output is the
    robustness of the current specification, $r(\phi_{\mathbf{w}},\tau)$, for
    that trajectory. The neural network is trained to minimize a regularized classification
    loss (discussed in \Cref{subsec:simplification-process}), given a labeled
    dataset $\mathcal{T}$.

    \begin{algorithm}[tb]
        \small
        \color{black}
        \caption{\color{black}Explanation Generation}
        \label{alg:expl}
        \color{black}
        \begin{algorithmic}
            [1] \Require Dataset $\mathcal{T}$, predicates
            $\Psi = \Set{\psi_i }_{i=1:N_\mathrm{AP}}$, similarity threshold $S_{\mathrm{th}}$,
            number of pruning iterations $N_{\mathrm{pr}}$, number of weights to
            prune $N_{\mathbf{w}}$, loss function $\tilde{\mathcal{L}}.$
            \Ensure Explanation $\phi$ \State Evaluate robustness of
            trajectories in $\mathcal{T}$ using $\Psi$. \State Filter out predicates
            whose similarities are above $S_{\mathrm{th}}$. \State Initialize
            the neural network $f_{\mathbf{w}}$ with parameters $\mathbf{w}$. \State
            Optimize $\mathbf{w}$ to minimize $\tilde{\mathcal{L}}$.
            \For{$i = 1:N_{\mathrm{pr}}$}
            \State Prune zeroed and smallest $N_{\mathbf{w}}$ parameters in $\mathbf{w}$.
            \State Re-normalize and optimize $\mathbf{w}$ to minimize $\tilde{\mathcal{L}}$.
            \If
            {only one non-zero weight in $\mathbf{w}^{\mathcal{F}}$/$\mathbf{w}^{\mathcal{G}}$}
            \textbf{break}. \EndIf \EndFor \State \Return $\phi$ generated from
            $\mathbf{w}$ and $\Psi$ using \Cref{eqn:nn_templ}.
        \end{algorithmic}
    \end{algorithm}

\begin{figure}[tb]
        \medskip
        \centering
        \includegraphics[width=0.9\linewidth]{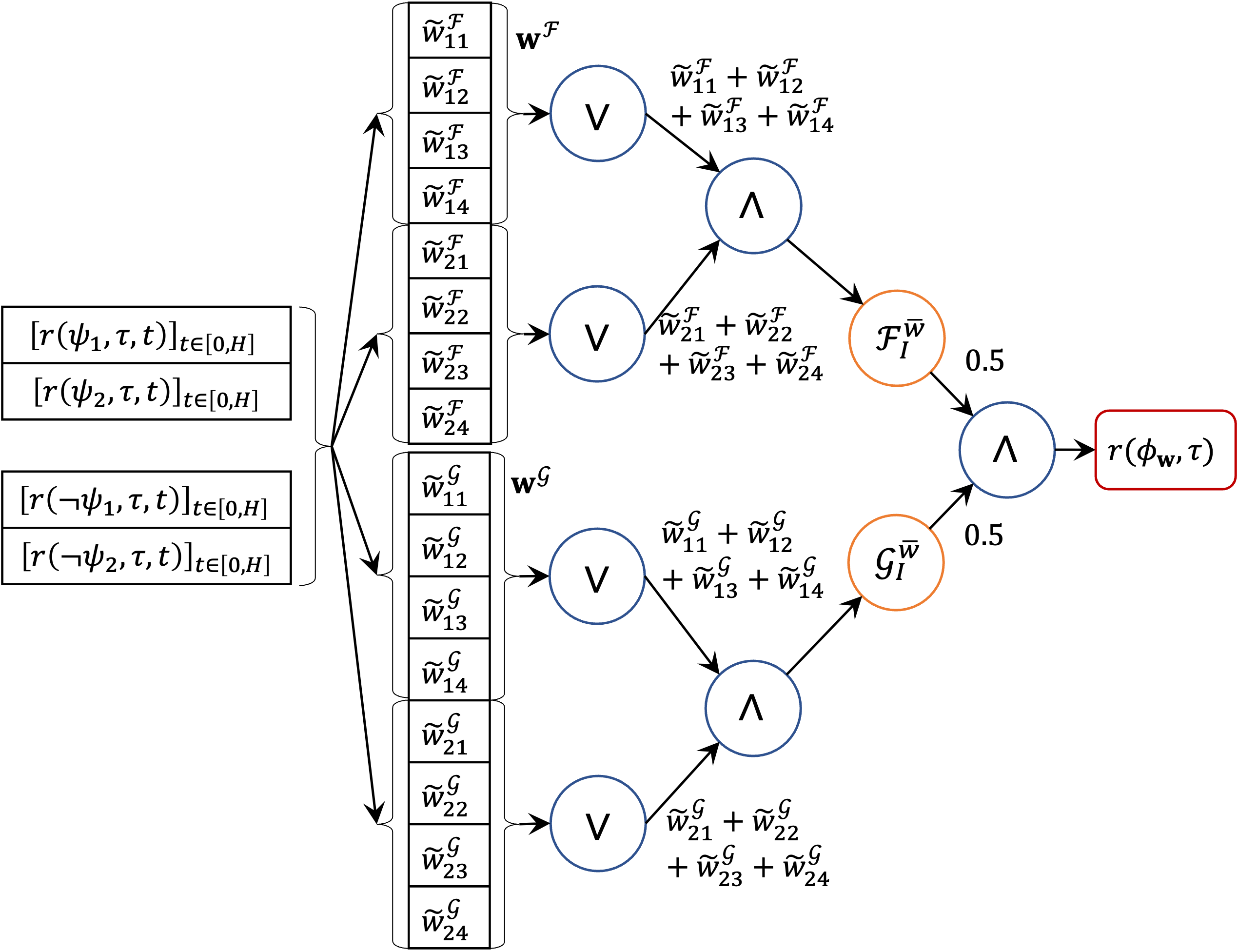}
        \caption{Example \tlnet architecture with
        two predicates, where $\phi_\mathbf{w}$ is the explanation generated from $\mathbf{w}$.}
        \label{fig:nn_arch} 
    \end{figure}

    We reduce the number of parameters to be optimized by distributing the weights
    in \Cref{eqn:nn_templ_standard}, resulting in a specification of the
    following form,
    \begin{align}
         & \phi_{\mathbf{w}}=\notag                                                                                                                                                                                                                                                                                           \\
         & 0.5\mathcal{F}_{I}^{\bar{w}}\left[\bigwedge_{i=1}^{N_\mathrm{AP}}\bigvee_{j=1}^{2N_\mathrm{AP}}\tilde{w}_{ij}^{\mathcal{F}}\psi_{j}\right] \wedge 0.5\mathcal{G}_{I}^{\bar{w}}\left[\bigwedge_{i=1}^{N_\mathrm{AP}}\bigvee_{j=1}^{2N_\mathrm{AP}}\tilde{w}_{ij}^{\mathcal{G}}\psi_{j}\right], \label{eqn:nn_templ}
    \end{align}
    where $\mathbf{w}=\mathbf{w}^{\mathcal{F}}\mathbin\Vert \mathbf{w}^{\mathcal{G}}$,
    $\mathbf{w}^{\mathcal{F}}= [\tilde{w}_{ij}^{\mathcal{F}}]_{i=1:N_\mathrm{AP},
    j=1:2N_\mathrm{AP}}$, $\mathbf{w}^{\mathcal{G}}= [\tilde{w}_{ij}^{\mathcal{G}}
    ]_{i=1:N_\mathrm{AP}, j=1:2N_\mathrm{AP}}$,
    $\tilde{w}_{ij}^{\mathcal{F}}= w_{i}^{\mathcal{F}}w_{ij}^{\mathcal{F}}$, and
    $\tilde{w}_{ij}^{\mathcal{G}}= w_{i}^{\mathcal{G}}w_{ij}^{\mathcal{G}}$. 
    \rev{This weight distribution reduces the number of optimized parameters from $2N_\mathrm{AP}+4N_\mathrm{AP}^{2}$ to $4N_\mathrm{AP}^{2}$.}
    We justify this weight distribution operation in \Cref{subsec:theory}.
    \Cref{fig:nn_arch} shows a \tlnet architecture for a dataset containing two predicates.

    \subsection{Simplification Process}
    \label{subsec:simplification-process}
    A key component of \tlnet is its simplification process, which aims to produce
    more human-interpretable wSTL explanations while preserving classification performance.
    This process consists of three steps: a predicate filter, loss
    regularization, and weight pruning.

    \subsubsection{Predicate Filter}
    As a pre-processing step, we implement a predicate filter that eliminates
    predicates that exhibit similar behavior across both positive and negative
    trajectory sets, as they contribute little to discriminating between those sets
    and may introduce unnecessary complexity. Given an arbitrary dataset of
    trajectories $\mathcal{D}$ and an atomic predicate $\psi$, we define the \emph{robustness
    distribution} for $\psi$ as,
    \begin{align}
         & d(\mathcal{D}, \psi) = \frac{1}{|\mathcal{D}|}\left[\abs{\mathcal{D}_1}, \abs{\mathcal{D}_2}, \abs{\mathcal{D}_3}\right],\\
        &\mathcal{D}_{1} := \set{\tau | r(\psi,\tau,t)\geq0 \quad \forall t\in [0,H], \tau \in \mathcal{D}}, \\
        &\mathcal{D}_{2} := \mathcal{D}\setminus (\mathcal{D}_{1}\cup \mathcal{D}_{3}),                      \\
        &\mathcal{D}_{3} := \set{\tau | r(\psi,\tau,t)<0 \quad \forall t\in [0,H], \tau \in \mathcal{D} }.
    \end{align}
    That is, $\mathcal{D}_{1}$ is the set of trajectories where $\psi$ is always
    satisfied, $\mathcal{D}_{2}$ is the set where $\psi$ is satisfied at least once,
    and $\mathcal{D}_{3}$ is the set where $\psi$ is never satisfied.

    Now, let $\mathcal{T}_{p}= \{(\tau, l) | l = 1, (\tau, l)\in\mathcal{T}\}$ be
    the positive set of trajectories in $\mathcal{T}$ and $\mathcal{T}_{n}= \mathcal{T}
    \setminus \mathcal{T}_{p}$ be the negative set. For each predicate
    $\psi \in \Psi$, we compute $d(\mathcal{T}_{p},\psi)$ and $d(\mathcal{T}
    _{n},\psi)$, and then calculate the cosine similarity $S$ between those
    distributions.
    We then remove predicates from the set $\Psi$ (and their negations from
    $\Psi_{\neg}$) which have a cosine similarity $S \ge S_{\mathrm{th}}$, where
    $S_{\mathrm{th}}$ is a user-defined threshold.

    \subsubsection{Loss Regularization}
    The regularization step uses two regularizers to improve the conciseness of
    inferred explanations. The first regularizer, a \textit{temporal clause
    regularizer}, $R_{\mathrm{T}}$, is defined as,
    \begin{multline}
        R_{\mathrm{T}}(\mathbf{w}) := \sum_{j=1}^{N_\mathrm{AP}}\max\left(\sum_{i=1}
        ^{N_\mathrm{AP}}\tilde{w}_{ij}^{\mathcal{F}},\sum_{i=1}^{N_\mathrm{AP}}\tilde
        {w}_{i(j+N_\mathrm{AP})}^{\mathcal{F}}\right)\\
        \max\left(\sum_{i=1}^{N_\mathrm{AP}}\tilde{w}_{ij}^{\mathcal{G}},\sum_{i=1}
        ^{N_\mathrm{AP}}\tilde{w}_{i(j+N_\mathrm{AP})}^{\mathcal{G}}\right).
    \end{multline}
    Intuitively, this regularizer encourages the task and constraint clauses to
    contain different atomic predicates, \rev{and is independent of clause sparsity.}

    The second regularizer, a \textit{disjunctive clause regularizer}, $R_{\mathrm{D}}$,
    is defined as,
    \begin{equation}
        R_{\mathrm{D}}(\mathbf{w}^{\mathcal{O}}) := \sum_{i=1}^{N_\mathrm{AP}-1}\sum
        _{j=i+1}^{N_\mathrm{AP}}\sum_{k=1}^{2N_\mathrm{AP}}\tilde{w}^{\mathcal{O}}
        _{ik}\tilde{w}^{\mathcal{O}}_{jk},
    \end{equation}
    where $\mathbf{w}^{\mathcal{O}}\in \{\mathbf{w}^{\mathcal{F}}, \mathbf{w}^{\mathcal{G}}
    \}$ defines the weights within one of our temporal operators and $\tilde{w}^{\mathcal{O}}
    _{ij}, \tilde{w}^{\mathcal{O}}_{jk}\in \mathbf{w}^{\mathcal{O}}$.
    Intuitively, this regularizer encourages disjunctive clauses within a temporal clause to contain different predicates, \rev{and is again independent of clause sparsity.}

    Given $(\tau, l) \in \mathcal{T}$, our overall loss, $\Tilde{\mathcal{L}}$, is
    then,
    \begin{align}
        \tilde{\mathcal{L}}(\tau, l, \mathbf{w}) = \mathcal{L}(\tau, l, \mathbf{w}) + & \lambda_{R_\mathrm{T}}R_{\mathrm{T}}(\mathbf{w})\notag                                                                          \\
        +                                                                             & \lambda_{R_\mathrm{D}}[R_{\mathrm{D}}(\mathbf{w}^{\mathcal{F}})+R_{\mathrm{D}}(\mathbf{w}^{\mathcal{G}})], \label{eqn:loss_all}
    \end{align}
    where $\mathcal{L}$ is a classification loss and
    $\lambda_{R_\mathrm{T}}, \lambda_{R_\mathrm{D}}\in \mathbb{R}_{\geq 0}$ are
    regularizer hyperparameters. We use the classification loss from \cite{yan_stone_2021},
    \begin{align}
        \mathcal{L}(\tau, l, \mathbf{w}) = \exp(-\zeta l r(\phi_{\mathbf{w}}, \tau)),
    \end{align}
    where $\zeta \in \mathbb{R}_{>0}$ is a hyperparameter.

    \subsubsection{Weight Pruning}
    The weight pruning step further simplifies the currently inferred
    explanation by removing unnecessary weights.
    First, all weights with a value of zero are removed from the network. Then,
    among the remaining weights, the $N_{\mathrm{w}}\in \mathbb{Z}_{>0}$ smallest
    weights---where $N_{\mathrm{w}}$ is a user-specified hyperparameter---are removed
    to eliminate the least contributing parts of the specification. The
    remaining weights are then normalized. The pruning process is performed $N_\mathrm{pr}\in \mathbb{Z}_{>0}$ times and terminates if there is only one non-zero weight in $\mathbf{w}^{\mathcal{F}}$ or $\mathbf{w}^{\mathcal{G}}$ to ensure the desired task-constraint structure.
    \rev{Empirically, increasing $N_\mathrm{w}$ leads to training instability, since pruning changes the computation graph structure of the neural network. Decreasing $N_{\mathrm{pr}}$ can lead to unnecessarily complex explanations.}

    \subsection{Trajectory Sampling}
    We obtain positive trajectories in $\mathcal{T}$ by executing the target policy
    in the environment. These trajectories can also be sampled from existing
    datasets. We obtain negative trajectories in $\mathcal{T}$ by executing a random-walk
    policy in the environment.
    \subsection{Explainability Evaluation Metrics}
    Beyond classification performance, \tlnet also emphasizes the human interpretability
    of inferred explanations.
    We quantify this emphasis by introducing three explainability metrics---\emph{conciseness},
    \emph{consistency}, and \emph{strictness}.

    Let $\phi$ be a wSTL explanation we want to evaluate and $N$ be the number
    of temporal clauses that we expect to appear in the explanation. For example,
    in this work, we expect the inferred explanations to have one $\mathcal{F}$ clause
    and one $\mathcal{G}$ clause (thus $N=2$) because we assume there should be a
    task component and a constraint component. Now, let $\phi_{n}$ be the $n$-th
    expected temporal clause in $\phi$, where $\phi_{n}= \emptyset$ if that
    clause does not exist in $\phi$. Let the (unweighted) STL counterpart of $\phi
    _{n}$ be $\hat{\phi}_{n}$, where $n \in \{1,\dots N\}$. For example, in our
    work, if $\phi = \mathcal{G}(0.1\psi_{1}\wedge 0.9 \psi_{2})$, then
    $\phi_{1}= \emptyset$, $\hat{\phi}_{1}= \emptyset$,
    $\phi_{2}= \mathcal{G}(0.1\psi_{1}\wedge 0.9 \psi_{2})$, and $\hat{\phi}_{2}=
    \mathcal{G}(\psi_{1}\wedge \psi_{2})$.

    \subsubsection{Conciseness}
    We define the \emph{conciseness} of $\phi$ as,
    \begin{equation*}
        \mathrm{Conciseness}:= \frac{1}{N}\sum_{n=1}^{N}
        \begin{cases}
            \frac{1}{m_{n}} & \text{if $m_{n}\neq 0$}, \\
            0               & \text{otherwise},
        \end{cases}
    \end{equation*}
    where $m_{n}$ is the number of predicates present in $n$-th expected temporal
    clause $\phi_{n}$. Conciseness ranges from 0 to 1, with higher scores implying
    a shorter explanation with at least one predicate inside each expected temporal
    clause.

    \subsubsection{Consistency}
    Let $\Phi = [ \phi^{k}]_{k=1:K}$ be a sequence of $K$ explanations for the same
    policy (e.g., from different training seeds). Then, let $\Phi_{n}= [\phi_{n}^{k}
    ]_{k=1:K}$ be the sequence of the $n$-th expected temporal clauses in $\Phi$,
    where $\phi_{n}^{k}$ is the $n$-th expected temporal clause for $\phi^{k}$. Let
    the STL counterpart of $\Phi_{n}$ be $\hat{\Phi}_{n}$.
    Then, let the set of unique elements of $\hat{\Phi}_{n}$ be $\Sigma_{n}= \{\hat
    {\phi}_{n}\in \hat{\Phi}_{n}\}$.
    We define the \textit{consistency} of $\Phi$ as,
    \begin{align*}
         & \mathrm{Consistency}:=                                                                                                                                                                                                                 \\
         & \frac{1}{N}\sum_{n=1}^{N}\begin{cases}\max_{\sigma_n \in \Sigma_n}\sum_{\hat{\phi}_n \in \hat{\Phi}_n}\frac{\mathbb{I}(\sigma_n=\hat{\phi}_n)}{K\abs{\Sigma_n}}&\text{if $\Sigma_{n}\neq \emptyset$},\\ 0&\text{otherwise}.\end{cases}
    \end{align*}
    Consistency ranges from 0 to 1, with 
    higher scores indicating greater agreement in clause structures across
    explanations, reflecting structural stability in learned explanations.

    \subsubsection{Strictness}
    Finally, let $P$ be the number of potentially present predicates in $\phi$ (e.g.,
    $P = 2N_{\mathrm{AP}}$ in \tlnet). Denote the numbers of conjunctions and disjunctions
    in $\phi_{n}$ as $C_{n}$ and $D_{n}$, respectively. We define the \textit{strictness}
    of $\phi$ as,
    \begin{equation*}
        \mathrm{Strictness}:= \frac{1}{N}\sum_{n=1}^{N}
        \begin{cases}
            \frac{1}{P - C_{n}+ D_{n}} & \text{if $\phi_{n}\neq \emptyset$}, \\
            0                          & \text{otherwise.}
        \end{cases}
    \end{equation*}
    Strictness ranges from 0 to 1, with higher scores implying stricter logical
    expressions by rewarding conjunctions and penalizing disjunctions. \rev{For example, given a temporal clause $\phi_{0}= \mathcal{F}[(0.5 \psi_{0}\vee 0.25 \psi_{1}) \wedge 0.25\psi_{2}]$ and $P = 6$, we get $C_{0}= 1$ and $D_{0}= 1$, with strictness of 0.167 for $\phi_{0}$.}


    \subsection{Weight Distribution in wSTL}
    \label{subsec:theory}
    A key assumption in \tlnet is that the weights can be distributed across
    Boolean and temporal clauses.
    We now prove that such a distribution is theoretically sound.
    The normalized weights of a wSTL formula in CNF can be syntactically distributed
    across Boolean operators as follows.\footnote{
    \rev{ 
        Syntactic weight distributivity also holds for temporal operators by distributing operator weights to the normalized interval weights followed by distributions to the inner Boolean structures. 
        }
    }
    \begin{thm}
        [Syntactic Weight Distributivity of Boolean Operators] \label{thm:bool_dist}
        A normalized wSTL formula $\phi$ in CNF can be represented as follows,
        \begin{align}
            \phi = \bigwedge_{i=1}^{n}w_{i}\bigvee_{j=1}^{m_i}w_{ij}(\neg)\psi_{ij}\equiv \bigwedge_{i=1}^{n}\bigvee_{j=1}^{m_i}\tilde{w}_{ij}(\neg)\psi_{ij},
        \end{align}
        where $n, m_{i}\in\mathbb{Z}_{>0}$, $w_{i}, w_{ij}\in\mathbb{R}_{>0}$,
        $\tilde{w}_{ij}:=w_{i}w_{ij}$, $\sum_{i=1}^{n}w_{i}= \sum_{j=1}^{m_i}w_{ij}
        = 1$, and $\psi_{ij}$ are predicates. Furthermore, the weights
        $\tilde{w}_{ij}$ satisfy
        $\sum_{i=1}^{n}\sum_{j=1}^{m_i}\tilde{w}_{ij}= 1$.
        \begin{proof}
            \begin{align*}
                \{\text{LHS}\} & = w_{1}(w_{11}(\neg)\psi_{11}\vee ... \vee w_{1m_1}(\neg)\psi_{1m_1})            \\
                ...            & \wedge w_{n}(w_{n1}(\neg)\psi_{n1}\vee ... \vee w_{nm_n}(\neg)\psi_{nm_n})       \\
                               & = (w_{1}w_{11}(\neg)\psi_{11}\vee ... \vee w_{1}w_{1m_1}(\neg)\psi_{1m_1})       \\
                ...            & \wedge (w_{n}w_{n1}(\neg)\psi_{n1}\vee ... \vee w_{n}w_{nm_n}(\neg)\psi_{nm_n})  \\
                               & =\bigwedge_{i=1}^{n}\bigvee_{j=1}^{m_i}w_{i}w_{ij}(\neg)\psi_{j}=\{\text{RHS}\}.
            \end{align*}
        \end{proof}
    \end{thm}

    \section{Results}

    We perform a series of empirical experiments to compare our method to relevant
    baselines for robotic tasks. We consider three baselines for inference of
    wSTL explanations, each adapted to model our normalized wSTL structures (given
    by \Cref{eqn:nn_templ}):
    \begin{outline}
        \1 \emph{NN-TLI} \cite{li_learning_2023}: this method implements a greedy
        pruning strategy that independently removes network weights that do not affect
        classification accuracy after optimization. We removed their weight-rounding
        process (to convert to STL) to consider the structure in
        \Cref{eqn:nn_templ} and used the aggregation functions given in
        \Cref{eqn:agg_and,eqn:agg_or,eqn:agg_g,eqn:agg_f}, since our empirical results showed more training stability
        with those functions.
        \1 \emph{STONE (top-3, top-5)} \cite{yan_stone_2021,yan_neuro-symbolic_2022}: this method retains the 3 or 5
        largest network weights after training. 
        \1 \rev{\emph{DT (STL)} \cite{bombara_offline_2021}: this method uses decision trees to mine STL (not wSTL) explanations.}
    \end{outline}
    The first two baselines, NN-TLI and STONE (top-3, top-5), were implemented using the same neural network architecture as
    \tlnet to ensure a consistent basis for comparison. 
    For each policy we explain, we collected 500 positive trajectories using the trained policy and 500
    negative trajectories using a random-walk policy. Each dataset was split 80\%
    and 20\% for training and evaluation, respectively. \rev{Hyperparameter values are summarized in \Cref{tab:hyperparams}.}
    Our code\rev{, including videos of target policies,} is available online\footnote{\finrev{https://github.com/LIRA-illinois/tlnet}}.


     \begin{figure}[tb]
        \centering
        \hfill
        \begin{tabular}{@{}c@{}}
            \subfloat[CtF environment.\label{fig:game_field}]{\includegraphics[width=0.39\linewidth]{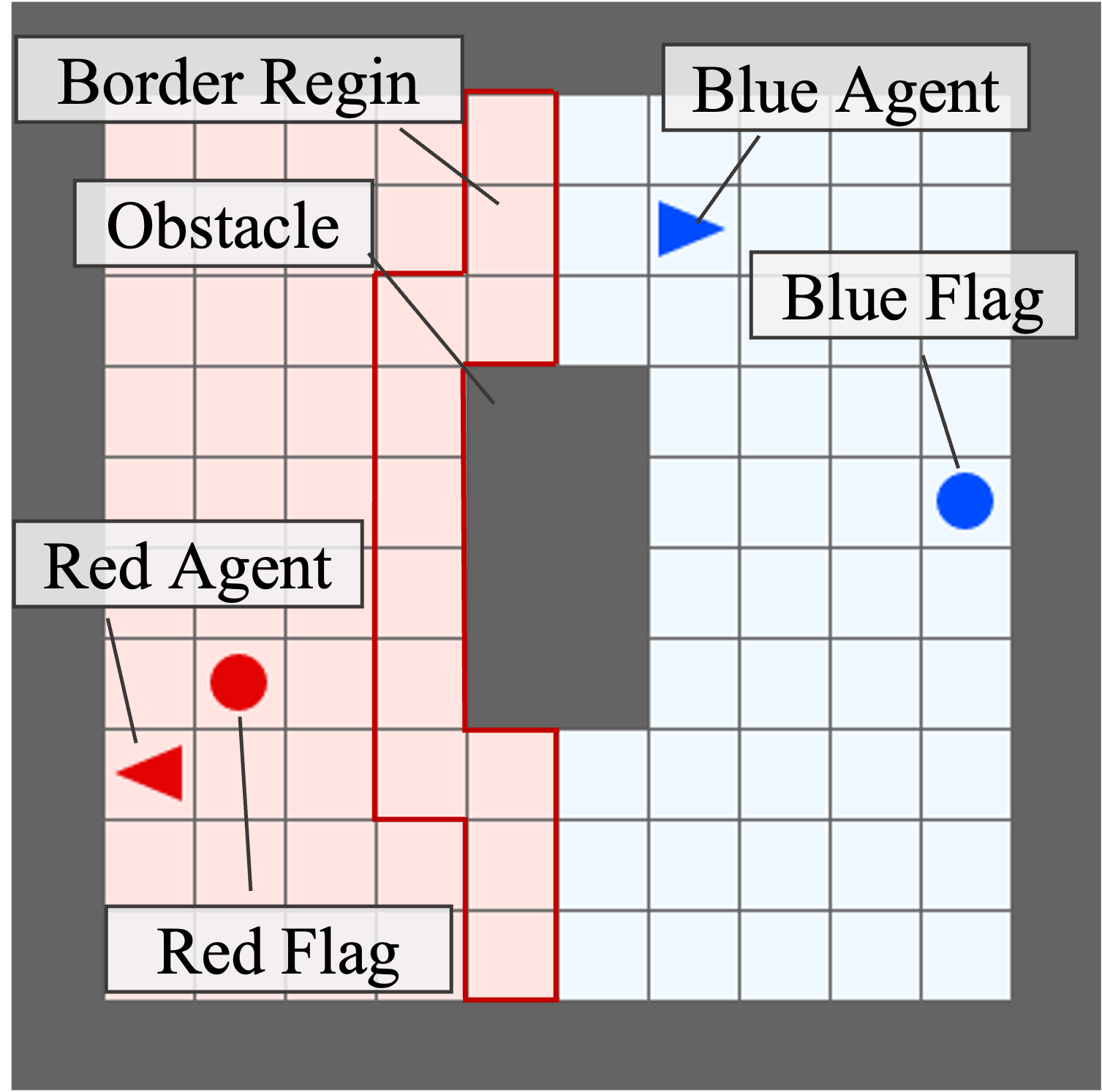}}
        \end{tabular}
        \hfill
        \begin{tabular}{@{}c@{}}
            \subfloat[Obstructed fetch push.\label{fig:fetch_env}]{\includegraphics[width=0.28\linewidth]{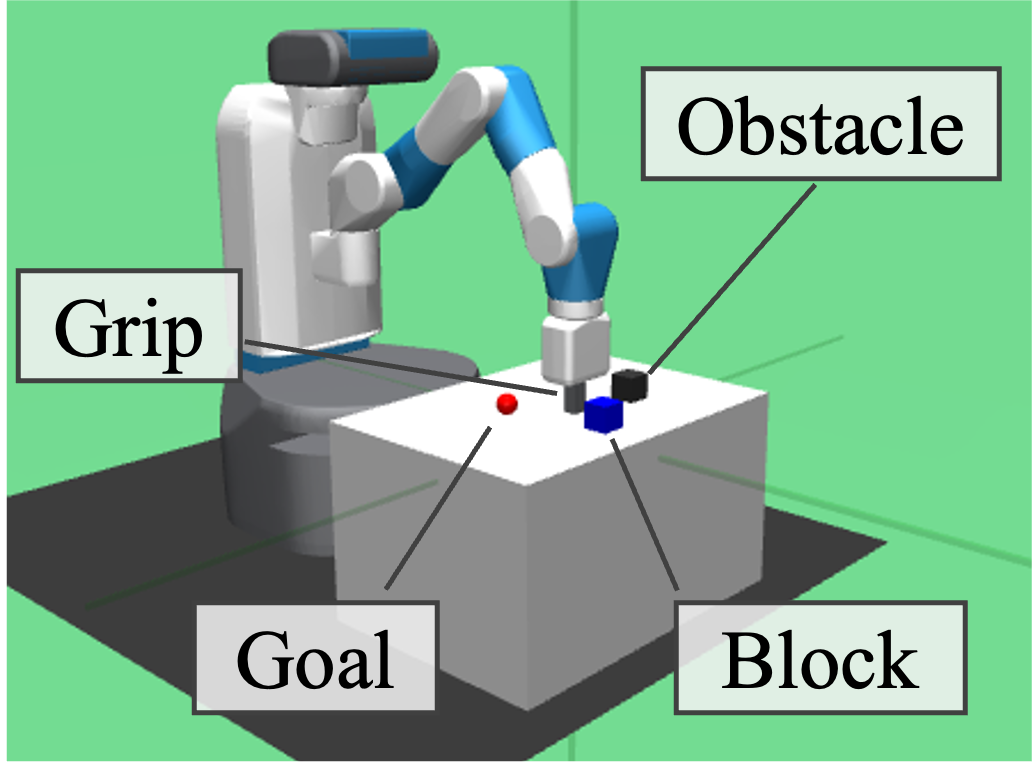}} \\
        \end{tabular}
        \hfill
        \begin{tabular}{@{}c@{}}
            \subfloat[Robot navigation.\label{fig:robot_env}]{\includegraphics[width=0.26\linewidth]{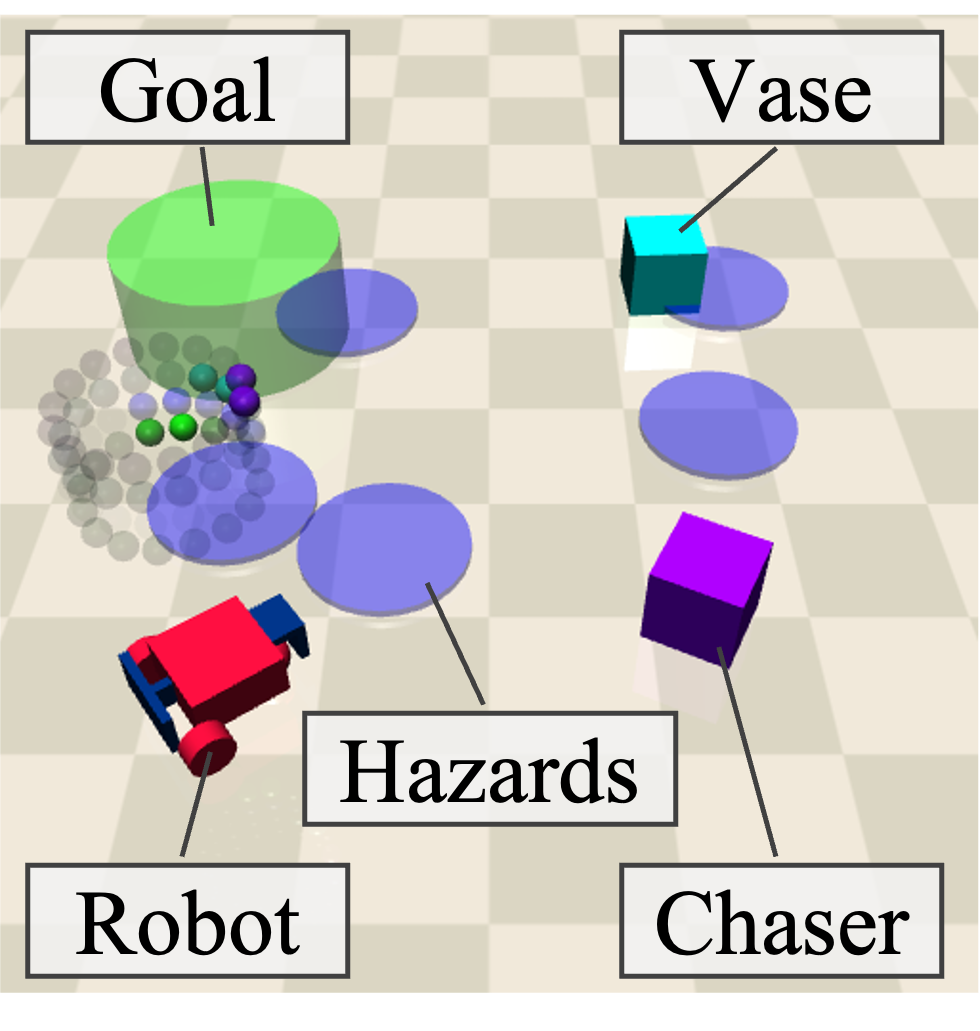}}
        \end{tabular}
        \hfill
        \caption{Initial states of our test environments with relevant objects shown.}
        \label{fig:envs}
    \end{figure}

    \begin{table}[tb]
        \centering
        \medskip
        \caption{\rev{Hyperparameter values and their ranges.}}
        \tabcolsep = 3pt \rowcolors{2}{gray!25}{white}
        \begin{tabular}{l|llllllll}
            Environments  & $\sigma$          & $\zeta$           & $S_{\mathrm{th}}$ & $\lambda_{R_\mathrm{T}}$ & $\lambda_{R_\mathrm{D}}$ & $N_{\mathrm{pr}}$ & $N_{\mathrm{w}}$  & $H$               \\
            \hline
            CtF           & 0.5               & 1.0               & 0.99              & 0.01                     & 0.1                      & 34                & 1                 & 100               \\
            Fetch Push    & 0.5               & 1.0               & 0.99              & 0.01                     & 0.1                      & 20                & 1                 & 200               \\
            Robot Nav.    & 0.5               & 1.0               & 0.99              & 0.01                     & 0.1                      & 28                & 1                 & 400               \\
            \hline
            Range ($\in$) & $\mathbb{R}_{>0}$ & $\mathbb{R}_{>0}$ & [-1, 1]           & $\mathbb{R}_{\ge0}$      & $\mathbb{R}_{\ge0}$      & $\mathbb{Z}_{>0}$ & $\mathbb{Z}_{>0}$ & $\mathbb{Z}_{>0}$
        \end{tabular}
        \label{tab:hyperparams}
    \end{table}

    \subsection{Test Environments}

    We consider the three environments shown in \Cref{fig:envs}, which include discrete
    and continuous state-action spaces, navigation and manipulation tasks, and varied
    predicates.

    \subsubsection{Capture-the-flag (CtF)}
    CtF is a discrete grid-world environment with adversarial dynamics where two
    agents compete to capture each other’s flags, based on
    \cite{yuasa_generating_2024}. If both agents occupy adjacent cells in a given
    territory, the defending agent eliminates the other with 75\% probability.
    We considered blue agent policies optimized for five scenarios. The first four
    considered different red agent heuristic policies (\textit{capture}, \textit{fight},
    \textit{patrol}, \textit{mix}) with the same reward function (capture: +1,
    kill: +0.5, being captured: $-1$, timestep penalty: $-0.01$). The fifth
    scenario (\textit{capture 0}) used a \rev{\textit{capture}} red agent with
    the reward function modified to remove the kill reward (kill: 0). The red
    agent policies are defined as follows: \textit{capture} pursues the flag; \textit{fight}
    pursues the blue agent; \textit{patrol} defends a boundary region; and \textit{mix}
    \rev{uses \textit{fight} when the blue agent is in the red territory and \textit{capture} otherwise.}
    Red agents perform a random action 25\% of the time.
    We defined eight atomic predicates based on the defeated (``df'') status
    $\{1, -1\}$ of the red agent and Euclidean distances among the blue agent (``ba''),
    blue flag (``bf''), blue territory (``bt''), red agent (``ra''), red flag (``rf''),
    and obstacle (``ob''): $\psi_{\mathrm{ba,bt}}$, $\psi_{\mathrm{ba,ob}}$,
    $\psi_{\mathrm{ba,ra}}$, $\psi_{\mathrm{ba,rf}}$, $\psi_{\mathrm{ra,bf}}$, $\psi
    _{\mathrm{ra,bt}}$, $\psi_{\mathrm{ra,df}}$, $\psi_{\mathrm{ra,ob}}$.

    \subsubsection{Fetch Push}
    We also consider a continuous control task based on the Fetch mobile manipulator
    \cite{gymnasium_robotics2023github}. The robot must push a blue block (``b'')
    to a red target (``t'') using its grip (``g''), without disturbing a black
    obstacle (``o''). The state and action spaces are continuous ($s \in \mathbb{R}
    ^{55}, a \in \mathbb{R}^{4}$). Eight predicates were defined based on the distances
    between the grip, block, target, and obstacle, including drop-related (``d'')
    conditions: $\psi_{\mathrm{gb}}$, $\psi_{\mathrm{gt}}$, $\psi_{\mathrm{go}}$,
    $\psi_{\mathrm{bt}}$, $\psi_{\mathrm{bo}}$, $\psi_{\mathrm{ot}}$,
    $\psi_{\mathrm{bd}}$, $\psi_{\mathrm{od}}$.

    \subsubsection{Robot Navigation}
    Finally, we extended the CarGoal-v0 environment \cite{ji_safety_2023} to
    create a more complex adversarial navigation task. The ego robot (``e'') must
    reach a green goal (``g'') while avoiding a violet chaser (``c'') moving with
    a fixed speed ($v_{c}= 7.5 \times 10^{-4}$). Static hazards (``h'') and a
    movable vase (``v'') are present but not penalized. The state and action dimensions
    are $s \in \mathbb{R}^{88}$ and $a \in \mathbb{R}^{2}$, respectively. Ten distance-based
    predicates were defined, involving Euclidean distances between the robot, the
    chaser, the goal, and the hazards: $\psi_{\mathrm{ec}}$, $\psi_{\mathrm{eg}}$,
    $\psi_{\mathrm{eh}}$, $\psi_{\mathrm{ev}}$, $\psi_{\mathrm{gc}}$, $\psi_{\mathrm{gh}}$,
    $\psi_{\mathrm{gv}}$, $\psi_{\mathrm{hc}}$, $\psi_{\mathrm{hv}}$, $\psi_{\mathrm{vc}}$.

    We included predicates irrelevant to task success in each environment (e.g.,
    $\psi_{\mathrm{ra,ob}}$, $\psi_{\mathrm{ot}}$, $\psi_{\mathrm{hv}}$) to evaluate
    whether the considered methods could eliminate them from their explanations.
    \rev{
    All predicates are distance-based (e.g., $d_\mathrm{ba,ra} < 0.5$ ) except for $\psi_\mathrm{ra,df}$ (binary).
    }
    \rev{Average \tlnet optimization times were 170.8 s (CtF), 118.3 s (Fetch Push), and 116.1 s (Robot Navigation).\footnote{\rev{With an Intel Xeon Silver 4314 CPU and NVIDIA RTX A4000 GPU.}}
        }

    \subsection{Policy Optimization}

    \tlnet is a general framework to generate explanations of robot policies, though,
    we focused on policies optimized by deep reinforcement learning algorithms
    in this work. We used Stable-Baselines3 \cite{raffin_stable-baselines3_2021}
    for training all policies. For the CtF and navigation environments, we used PPO \cite{schulman_proximal_2017}; 
    for the Fetch Push task, we used SAC \cite{haarnoja_soft_2018} with
    hindsight experience replay (HER) \cite{andrychowicz_hindsight_2017}.
    Hyperparameters were taken from prior work
    \cite{yuasa_generating_2024,raffin_rl_2020,ji_safety_2023}.
    For each policy, we performed 10 independent optimization runs with random
    initialization and collected classification accuracy and explainability
    metrics.

     \begin{table*}[tb]
        \centering
        \medskip
        \caption{Comparison of the Most Frequently Inferred Explanations \rev{with Frequencies (\%).}}
        \tabcolsep = 0.55pt 
        \rowcolors{2}{gray!25}{white}
        \begin{tabular}{l|lr|lr|lr|lr|lr}
            Scenario     & \multicolumn{2}{c|}{\tlnet (Ours)}  & \multicolumn{2}{c|}{NN-TLI}    &  \multicolumn{2}{c|}{STONE (top-3)}   &  \multicolumn{2}{c|}{STONE (top-5)}    & \multicolumn{2}{c}{\rev{DT}}   \\
            \hline
            \begin{tabular}{l}
                CtF \\ Capture
            \end{tabular}    
            & \begin{tabular}{l}
                $0.5\mathcal{F}_{I}^{\bar{w}}[1.0 \psi_{\mathrm{ba,rf}}] \wedge$ \\ $0.5\mathcal{G}_{I}^{\bar{w}}[0.30 \psi_{\mathrm{ba,rf}}\vee $\\ \qquad\quad $0.70 \neg\psi_{\mathrm{ra,bf}}]$ 
            \end{tabular} 
            & 60\% 
            &\begin{tabular}{l}
                $0.5\mathcal{F}_{I}^{\bar{w}}[0.26 \psi_{\mathrm{ba, rf}}\wedge (0.25 \psi_{\mathrm{ba, rf}}\vee 0.33$\\ $ \neg\psi_{\mathrm{ra, bt}}) \wedge(0.09 \neg\psi_{\mathrm{ba, bt}}\vee 0.07 \neg\psi_{\mathrm{ra, bt}})] \wedge$\\ $0.5\mathcal{G}_{I}^{\bar{w}}[0.36 \psi_{\mathrm{ba, rf}}\vee 0.64 \neg\psi_{\mathrm{ra, bf}}]$ 
            \end{tabular} 
            & 20\%
            &\begin{tabular}{l}
                $\mathcal{F}_{I}^{\bar{w}}[0.73  \neg$\\ $\psi_{\mathrm{ra,bt}}\wedge 0.$\\ $ 27 \neg\psi_{\mathrm{ba,bt}}]$
            \end{tabular} 
            & 50\%
            & \begin{tabular}{l}
                $\mathcal{F}_{I}^{\bar{w}}[0.83 \neg $\\ $ \psi_{\mathrm{ra,bt}}\wedge  0$\\ $.17 \psi_{\mathrm{ba,rf}}]$
            \end{tabular} 
            & 60\% 
            & \begin{tabular}{l}
                $\mathcal{F}_{I}[\psi_{\mathrm{ba,rf}}  $\\  $\vee \psi_{\mathrm{ra,bf}} \vee $\\  $ \neg\psi_{\mathrm{ba,bt}}]$
            \end{tabular}
            & 100\%    \\

            \begin{tabular}{l}
                CtF \\  Fight
            \end{tabular}      
            & \begin{tabular}{l}
                $0.5\mathcal{F}_{I}^{\bar{w}}[1.0 \psi_{\mathrm{ba,rf}}] \wedge$ \\ $0.5\mathcal{G}_{I}^{\bar{w}}[0.33 \psi_{\mathrm{ba,rf}}\vee$\\ \qquad\quad  $ 0.67 \neg\psi_{\mathrm{ba,ra}}]$
            \end{tabular} 
            & 70\%
            & \begin{tabular}{l}$0.5\mathcal{F}_{I}^{\bar{w}}[0.77 \psi_{\mathrm{ba, rf}}\wedge (0.15 \psi_{\mathrm{ba, rf}}\vee 0.08 \psi_{\mathrm{ra, bf}})] \wedge$\\ $0.5\mathcal{G}_{I}^{\bar{w}}[(0.15 \neg\psi_{\mathrm{ba, bt}}\vee 0.11 \neg\psi_{\mathrm{ra,df}}) \wedge 0.06 \psi_{\mathrm{ba, rf}}$\\ $\vee 0.04 \neg\psi_{\mathrm{ra,df}}) \wedge(0.06 \psi_{\mathrm{ba, rf}}\vee 0.06 \neg\psi_{\mathrm{ba, bt}}\vee$\\ $0.04 \neg\psi_{\mathrm{ra,df}}) \wedge (0.02 \psi_{\mathrm{ba,ra}}\vee 0.19 \psi_{\mathrm{ba, rf}}\vee$\\ $0.16 \neg\psi_{\mathrm{ba, bt}}\vee 0.11 \neg\psi_{\mathrm{ra,df}})]$
            \end{tabular} 
            & 70\%
            &\begin{tabular}{l}
                $\mathcal{F}_{I}^{\bar{w}}$\\$[1.0 \psi_{\mathrm{ba,rf}}]$
            \end{tabular}
            & 100\%
            & \begin{tabular}{l}
                $\mathcal{F}_{I}^{\bar{w}}$\\$ [1.0\psi_{\mathrm{ba,rf}}]$
            \end{tabular}
            & 100\%
            &\begin{tabular}{l}
                $\mathcal{F}_{I}[\psi_{\mathrm{ba,rf}}]$\\ $ \wedge $\\ $\mathcal{G}_{I}[\psi_{\mathrm{ba,bt}}]$
            \end{tabular}
            & 100\%         \\

            \begin{tabular}{l}
                CtF \\ Patrol
            \end{tabular}   & 
            \begin{tabular}{l}
                $0.5\mathcal{F}_{I}^{\bar{w}}[1.0 \psi_{\mathrm{ba,rf}}] \wedge$ \\ $0.5\mathcal{G}_{I}^{\bar{w}}[0.52 \psi_{\mathrm{ba,rf}}\vee$ \\ \qquad\quad $ 0.48 \neg\psi_{\mathrm{ra,bt}}]$
            \end{tabular} 
            & 70\%
            &\begin{tabular}{l}$0.5\mathcal{F}_{I}^{\bar{w}}[0.35 \psi_{\mathrm{ba, rf}}\wedge (0.09 \psi_{\mathrm{ba, rf}}\vee 0.55 \psi_{\mathrm{ra, df}})] \wedge$\\ $0.5\mathcal{G}_{I}^{\bar{w}}[(0.29 \psi_{\mathrm{ba, rf}}\vee 0.04 \psi_{\mathrm{ra, bf}}\vee 0.07 \neg\psi_{\mathrm{ra, df}}) \wedge$\\ $(0.30 \psi_{\mathrm{ba, rf}}\vee 0.19 \psi_{\mathrm{ra, bf}}\vee 0.02 \neg\psi_{\mathrm{ba, bt}}) \wedge$\\ $0.04 \psi_{\mathrm{ra, bf}}\wedge 0.05 \neg\psi_{\mathrm{ra, bf}}]$
            \end{tabular}
            & 30\%                                                                                                                    
            &\begin{tabular}{l}
                $\mathcal{F}_{I}^{\bar{w}}$\\$[1.0 \psi_{\mathrm{ra,df}}]$                                                                              
            \end{tabular}
            & 100\%
            &\begin{tabular}{l}
                $\mathcal{F}_{I}^{\bar{w}}$\\$[1.0 \psi_{\mathrm{ra,df}}]$
            \end{tabular}
            & 90\%
            & \begin{tabular}{l}
                $\mathcal{G}_{I}[\neg\psi_{\mathrm{ba,rf}}$\\ $ \wedge \psi_{\mathrm{ba,bt}}]$
            \end{tabular} 
            & 100\%      \\

            \begin{tabular}{l}
                CtF \\ Mix
            \end{tabular}    & 
            \begin{tabular}{l}
                $0.5\mathcal{F}_{I}^{\bar{w}}[1.0 \psi_{\mathrm{ba,rf}}] \wedge$ \\ $0.5\mathcal{G}_{I}^{\bar{w}}[0.44 \psi_{\mathrm{ba,rf}}\vee$ \\ \qquad\quad $ 0.56 \neg\psi_{\mathrm{ra,bf}}]$
            \end{tabular}  
            & 60\%
            & \begin{tabular}{l}
              $0.5\mathcal{F}_I^{\bar{w}}[1.0 \psi_\mathrm{ba, rf}] \wedge 0.5\mathcal{G}_I^{\bar{w}}[0.21 \neg\psi_\mathrm{ra, bf} \wedge 0.47 $\\
              $\psi_\mathrm{ba, rf}\wedge (0.04 \neg\psi_\mathrm{ra, bf} \vee 0.05 \neg\psi_\mathrm{ra, df}) \wedge (0.18 \psi$\\
              $ _\mathrm{ba, rf}\vee0.03 \psi_\mathrm{ra, df} \vee 0.04 \neg\psi_\mathrm{ra, bf} \vee 0.02 \neg\psi_\mathrm{ra, bt})]$
            \end{tabular}
            & 10\%
            &\begin{tabular}{l}
                $\mathcal{F}_{I}^{\bar{w}}$\\$[1.0 \psi_{\mathrm{ba,bt}}]$
            \end{tabular}
            & 60\%
            & \begin{tabular}{l}$\mathcal{F}_{I}^{\bar{w}}[0.48 \neg $\\ $\psi_{\mathrm{ba,bt}}\wedge 0$\\ $ .52\psi_{\mathrm{ba,rf}}]$\end{tabular} 
            & 40\%
            &\begin{tabular}{l}
                $\mathcal{F}_{I}[\neg\psi_{\mathrm{ba,rf}}$\\ $  \vee\neg\psi_{\mathrm{ba,ra}}]$\\$ \wedge  \mathcal{G}_{I}[\psi_{\mathrm{ba,rf}}]$
            \end{tabular}
            & 100\%   \\

            \begin{tabular}{l}
                CtF \\ Capture \\ 0
            \end{tabular} & 
            \begin{tabular}{l}
                $0.5\mathcal{F}_{I}^{\bar{w}}[1.0 \psi_{\mathrm{ba,rf}}] \wedge$ \\ $0.5\mathcal{G}_{I}^{\bar{w}}[0.31 \psi_{\mathrm{ba,rf}}\vee $ \\\qquad\quad  $ 0.69 \neg\psi_{\mathrm{ra,bf}}]$
            \end{tabular} 
            & 80\%
            & \begin{tabular}{l} 
             $0.5\mathcal{F}_I^{\bar{w}}[0.08 \psi_\mathrm{ba, rf}\wedge (0.40 \psi_\mathrm{ba, rf} \vee 0.31 \neg\psi_\mathrm{ra, bt}) \wedge$\\ 
             $(0.07 \neg\psi_\mathrm{ba, bt} \vee 0.14 \neg\psi_\mathrm{ra, bt})] \wedge 0.5\mathcal{G}_I^{\bar{w}}[(0.33 \psi_\mathrm{ba,}$\\
             $ _\mathrm{rf}\vee 0.58 \neg\psi_\mathrm{ra, bf}) \wedge (0.05 \psi_\mathrm{ba, rf} \vee 0.04 \neg\psi_\mathrm{ra, bt})]$
            \end{tabular} 
            & 20\%
            & \begin{tabular}{l}$\mathcal{F}_{I}^{\bar{w}}[0.67\neg $\\ $\psi_{\mathrm{ra,bt}} \wedge 0. $\\ $33\neg\psi_{\mathrm{ba,bt}}]$\end{tabular} 
            & 60\%
            & \begin{tabular}{l}$\mathcal{F}_{I}^{\bar{w}}[0.83 \neg $\\ $\psi_{\mathrm{ra,bt}}\wedge 0$\\$.17 \psi_{\mathrm{ba,rf}}]$\end{tabular} 
            & 40\%
            & \begin{tabular}{l}
                $\mathcal{F}_{I}[\psi_{\mathrm{ba,rf}}\vee $\\ $ \neg\psi_{\mathrm{ba,ra}}\vee$\\ $\neg\psi_{\mathrm{ba,bt}}]$
            \end{tabular}
            & 100\%\\

            \begin{tabular}{l}
                Fetch \\ Push
            \end{tabular}    
            & \begin{tabular}{l}
                $0.5\mathcal{F}_{I}^{\bar{w}}[1.0 \psi_{\mathrm{bt}}] \wedge0.5$ \\ $\mathcal{G}_{I}^{\bar{w}}[0.41 \psi_{\mathrm{gb}}\vee 0.59 \psi_{\mathrm{bt}}]$
            \end{tabular}              
            & 100\%
            & \begin{tabular}{l}$0.5\mathcal{F}_{I}^{\bar{w}}[0.74 \psi_{\mathrm{bt}}\wedge (0.10 \psi_{\mathrm{bt}}\vee 0.16 \psi_{\mathrm{od}})] \wedge$\\ $0.5\mathcal{G}_{I}^{\bar{w}}[1.0 \neg\psi_{\mathrm{bd}}]$\end{tabular} 
            & 40\%
            & \begin{tabular}{l}
                $\mathcal{G}_{I}^{\bar{w}}[1.0\psi_{\mathrm{bt}}]$
            \end{tabular}
            & 100\%
            & \begin{tabular}{l}
                $\mathcal{G}_{I}^{\bar{w}}[1.0\psi_{\mathrm{bt}}]$     
            \end{tabular}
            & 100\%                                                                
            &    $\mathcal{G}_{I}[\neg\psi_{\mathrm{og}}]$
            & 100\%      \\

            \begin{tabular}{l}
                Robot \\ Nav.
            \end{tabular}   
            &\begin{tabular}{l}
                $0.5\mathcal{F}_{I}^{\bar{w}}[1.0 \psi_{\mathrm{eg}}] \wedge$ \\ $ 0.5\mathcal{G}_{I}^{\bar{w}}[1.0 \neg\psi_{\mathrm{ec}}]$
            \end{tabular}                                                                            
            & 70\%
            & $\mathcal{F}_{I}^{\bar{w}}[1.0 \psi_{\mathrm{eg}}]$      
            & 70\%         
            & \begin{tabular}{l}
                $\mathcal{F}_{I}^{\bar{w}}[1.0 \psi_{\mathrm{eg}}]$
            \end{tabular} 
            & 100\%
            & \begin{tabular}{l}
                $\mathcal{F}_{I}^{\bar{w}}[1.0 \psi_{\mathrm{eg}}]$
            \end{tabular} 
            & 100\%
            &\begin{tabular}{l}
                $\mathcal{F}_{I}[\psi_{\mathrm{eg}} \vee$\\ $ \psi_{\mathrm{ec}}\vee \psi_{\mathrm{eh}}]$
            \end{tabular}
            & 100\%
        \end{tabular}
        \label{tab:baseline_qual}
    \end{table*}

    \begin{table*}
        [tb]
        \centering
        \caption{\rev{Comparison of Conciseness, Consistency, Strictness with standard deviation. The best results are bolded.}\tablefootnote{\rev{
        Statistical significance (Friedman test, Dunn-Bonferroni analysis) against the second best values indicated with * ($p<0.05$), ** ($p<0.01$).}}}
        \rowcolors{2}{gray!25}{white} \tabcolsep = 1.4pt
        \begin{tabular}{l|l|l|ll|l|c|c|cc|c|l|l|ll|l}
                               & \multicolumn{5}{c|}{\textbf{Conciseness}} & \multicolumn{5}{c|}{\textbf{Consistency}} & \multicolumn{5}{c}{\textbf{Strictness}} \\
            \rowcolor{gray!25} &                                           &                                           & \multicolumn{2}{c|}{STONE}             &                  &                  &                & \multicolumn{2}{c|}{STONE} & &      &       & \multicolumn{2}{c|}{STONE} & \\
            Scenario          & Ours                                      & NN-TLI                                    & top-3                                  & top-5            & \rev{DT}           & Ours           & NN-TLI                     & top-3 & top-5 & \rev{DT}     & Ours                             & NN-TLI                         & top-3            & top-5            & \rev{DT}\\
            \hline
            CtF Capt.        & $\textbf{0.41}_{\pm0.06}^{*}$                 & $0.21_{\pm0.04}$                          & $0.20_{\pm0.05}$                     & 0.22$_{\pm0.12}$ & 0.12$_{\pm0.00}$   & 0.33           & 0.07                       & 0.08  & 0.13 & \textbf{0.50} & \textbf{0.12}$_{\pm0.01}^{*}$    & 0.10$_{\pm0.01}$               & 0.07$_{\pm0.01}$ & 0.08$_{\pm0.02}$ & 0.04$_{\pm0.00}$ \\
            CtF Fight          & \textbf{0.39}$_{\pm0.04}^{*}$                 & 0.18$_{\pm0.05}$                          & 0.25$_{\pm0.00}$                   & 0.25$_{\pm0.00}$ & 0.12$_{\pm0.00}$   & 0.68           & 0.09                       & 0.50  & 0.50 & \textbf{1.00} & \textbf{0.12}$_{\pm0.00}^{*}$    & 0.09$_{\pm0.01}$               & 0.06$_{\pm0.00}$ & 0.06$_{\pm0.00}$ & \textbf{0.12}$_{\pm0.00}^{*}$   \\
            CtF Patrol         & \textbf{0.38}$_{\pm0.04}^{*}$                 & 0.18$_{\pm0.10}$                          & 0.25$_{\pm0.00}$                   & 0.28$_{\pm0.07}$ & 0.50$_{\pm0.00}$   & \textbf{0.63}  & 0.05                       & 0.50  & 0.53 & 0.50          & \textbf{0.12}$_{\pm0.00}^{*}$    & 0.09$_{\pm0.02}$               & 0.06$_{\pm0.00}$ & 0.07$_{\pm0.02}$ & 0.07$_{\pm0.00}$ \\
            CtF Mix        & \textbf{0.39}$_{\pm0.06}^{*}$                  & 0.12$_{\pm0.06}$                          & 0.21$_{\pm0.05}$                     & 0.16$_{\pm0.09}$ & 0.17$_{\pm0.00}$   & 0.27           & 0.06                       & 0.10  & 0.09 & \textbf{1.00} & \textbf{0.11}$_{\pm0.01}^{*}$    & 0.07$_{\pm0.01}$               & 0.06$_{\pm0.01}$ & 0.07$_{\pm0.02}$ & \textbf{0.11}$_{\pm0.01}^{*}$  \\
            CtF Capt.0        & \textbf{0.42}$_{\pm0.04}^{*}$                  & 0.21$_{\pm0.05}$                          & 0.20$_{\pm0.04}$                  & 0.30$_{\pm0.13}$ & 0.42$_{\pm0.00}$   & 0.45           & 0.11                       & 0.15  & 0.24 & \textbf{0.50} & \textbf{0.11}$_{\pm0.00}$        & \textbf{0.11}$_{\pm0.02}$      & 0.07$_{\pm0.00}$ & 0.10$_{\pm0.03}$ & 0.04$_{\pm0.00}$ \\
            Fetch Push         & \textbf{0.42}$_{\pm0.00}^{*}$                 & 0.31$_{\pm0.08}$                          & 0.25$_{\pm0.00}$                   & 0.25$_{\pm0.00}$ & 0.25$_{\pm0.00}$   & \textbf{1.00 } & 0.28                       & 0.50  & 0.50 & 0.50          & \textbf{0.11}$_{\pm0.00}^{**}$   & \textbf{0.11}$_{\pm0.02}^{**}$ & 0.06$_{\pm0.00}$ & 0.06$_{\pm0.00}$ & 0.06$_{\pm0.00}$ \\
            Robot Nav.        & \textbf{0.47}$_{\pm0.04}^{**}$                 & 0.22$_{\pm0.06}$                          & 0.25$_{\pm0.00}$                   & 0.25$_{\pm0.00}$ & 0.12$_{\pm0.00}$   & \textbf{0.68}  & 0.15                       & 0.50  & 0.50 & 0.50          & \textbf{0.10}$_{\pm0.00}^{**}$   & 0.05$_{\pm0.01}$               & 0.05$_{\pm0.00}$ & 0.05$_{\pm0.00}$ & 0.04$_{\pm0.00}$
        \end{tabular}
        \label{tab:baseline_quant_exp}
    \end{table*}

    \begin{table}[tb]
        \centering
        \caption{\rev{Comparison of Accuracy. The best results are bolded.}}
        \rowcolors{2}{gray!25}{white} \tabcolsep = 3.5pt
        \begin{tabular}{l|l|l|ll|l}
                        & \multicolumn{4}{c}{\textbf{Accuracy}}\\
            \rowcolor{gray!25}             &                            &                            & \multicolumn{2}{c|}{STONE}  & \\
            Scenario   & Ours                       & NN-TLI                     & top-3                      & top-5                     & \rev{DT}\\
            \hline
            CtF Capt. & $\textbf{0.95}_{\pm 0.02}$ & $\textbf{0.95}_{\pm 0.01}$ & $\textbf{0.95}_{\pm 0.01}$ & $0.94_{\pm0.02}$          & \textbf{0.95}$_{\pm0.00}$\\
            CtF Fight   & \textbf{0.94}$_{\pm0.02}$  & 0.92$_{\pm0.02}$           & \textbf{0.94}$_{\pm0.02}$  & 0.93$_{\pm0.01}$          & \textbf{0.94}$_{\pm0.00}$\\
            CtF Patrol  & \textbf{0.95}$_{\pm0.02}$  & 0.93$_{\pm0.02}$           & 0.94$_{\pm0.01}$           & 0.94$_{\pm0.02}$          & \textbf{0.95}$_{\pm0.00}$\\
            CtF Mix & 0.87$_{\pm0.01}$           & \textbf{0.89}$_{\pm0.04}$  & 0.88$_{\pm0.03}$           & 0.88$_{\pm0.05}$          & \textbf{0.89}$_{\pm0.00}$\\
            CtF Capt.0 & \textbf{0.95}$_{\pm0.02}$  & \textbf{0.95}$_{\pm0.01}$  & \textbf{0.95}$_{\pm0.01}$  & \textbf{0.95}$_{\pm0.02}$ & \textbf{0.95}$_{\pm0.00}$\\
            Fetch Push  & \textbf{0.99}$_{\pm0.01}^{**}$  & 0.97$_{\pm0.02}$           & 0.97$_{\pm0.01}$           & 0.97$_{\pm0.01}$     & \textbf{0.99}$_{\pm0.00}^{**}$\\
            Robot Nav. & 0.97$_{\pm0.01}$  & 0.96$_{\pm0.01}$           & 0.96$_{\pm0.01}$           & 0.96$_{\pm0.01}$          & \textbf{0.98}$_{\pm0.01}$
        \end{tabular}
        \label{tab:baseline_quant_ac}
    \end{table}

    \subsection{Experimental Results}

    \subsubsection{Benefits \& Limitations of \tlnet}
    \Cref{tab:baseline_qual} presents the most frequent explanation obtained by each
    method for each scenario. Across all scenarios, \tlnet consistently inferred
    both task and constraint clauses, while also producing concise and
    relatively comprehensible explanations.
    In comparison, NN-TLI typically yielded verbose explanations that are
    difficult to understand, while STONE often captured only one temporal clause,
    resulting in incomplete explanations.

    The explanations produced by \tlnet are also qualitatively reasonable. 
    For all CtF scenarios, \tlnet correctly inferred the task
    clause as $\psi_{\mathrm{ba,rf}}$ (\textit{``blue agent captures the red
    flag''})---no other method correctly inferred this task for every scenario. The
    inferred CtF constraints are also syntactically reasonable. For example, in the
    Capture scenario, the blue agent (global) constraint includes \textit{``red
    agent not on the blue flag''} because the \textit{capture} red policy
    only targets the blue flag, which requires the blue agent to learn to prevent
    the red agent from reaching the blue flag. Other scenarios, such as Fight and
    Patrol, did not include this predicate in the inferred constraint because
    those red policies do not target the blue flag. In comparison, the CtF
    constraints inferred by NN-TLI are generally too verbose to understand, while
    STONE did not infer any constraints.
    The DT (STL) baseline often produced a task- or constraint-only STL explanation, typically with semantically-inaccurate clauses.
    For Fetch Push, \tlnet also correctly inferred the task clause as
    $\psi_{\mathrm{bt}}$ (\textit{``block reaches the target''}), which no other
    method correctly inferred. While \tlnet inferred part of the expected constraint
    ($\psi_{\mathrm{gb}}, \psi_{\mathrm{bt}}$), it missed an expected constraint
    $\neg\psi_{\mathrm{go}}$ (``grip does not touch the obstacle''), likely due to
    insufficient occurrences of the grip touching the obstacle in the sampled data.
     NN-TLI also inferred part of the expected constraint,
    but also missed the $\neg\psi_{\mathrm{go}}$ constraint.
    Meanwhile, both
    variations of STONE inferred an incorrect constraint of $\psi_{\mathrm{bt}}$
    (\textit{``block reaches the target''}), since this is actually the task and
    not possible to always satisfy.

    For Robot Navigation, all methods correctly inferred the task clause but \tlnet
    was the only one to infer the correct constraint clause as $\neg\psi_{\mathrm{ec}}$
    (\textit{``the robot does not hit the chaser''}).

    As shown in \Cref{tab:baseline_quant_exp,tab:baseline_quant_ac}, \tlnet also
    outperformed all baselines in conciseness, consistency, and strictness, with
    marginal differences in classification accuracy, \rev{except for DT (STL) in consistency.}
    Across all scenarios,
    \tlnet achieved up to 1.9$\times$ higher conciseness, 2.6$\times$ higher consistency,
    and 2.0$\times$ higher strictness.
    These results support the qualitative findings above, validating the explainability
    of our method's outputs and the use of these explainability metrics.

    \subsubsection{Benefit of Including Weights}
    We now demonstrate the benefit of including weights within inferred explanations
    (e.g., using wSTL rather than STL \cite{li_learning_2023,bombara_offline_2021,aasi_classification_2022} or LTL
    \cite{gaglione_learning_2021})
    by examining how inferred weights differ across CtF scenarios.
    Consider the Capture and Mix explanations in \Cref{tab:baseline_qual}.
    Both \rev{\textit{capture}} and \rev{\textit{mix}} red policies target
    the blue flag, but the \rev{\textit{capture}} policy is more aggressive in doing
    so. The blue agent, in response, learns slightly different policies: a more defensive
    policy in Capture, prioritizing the protection of its flag, and a more aggressive
    policy in Mix, focusing more on capturing the red flag. This policy difference
    is reflected in the weights in the constraint clause,
    where the $\neg\psi_{\mathrm{ra,bf}}$ predicate receives a higher weight in
    the Capture explanation (0.70) than in the Mix explanation (0.56), due to
    the Capture policy's prioritization of flag protection. Conversely,
    the Mix explanation shows a higher weight for the $\psi_{\mathrm{ba,rf}}$
    predicate (0.44) than the Capture explanation (0.30), due to its prioritization of flag capture. Note that $\psi_{\mathrm{ba,rf}}$ appears
    in both the task and constraint clauses in these explanations. This
    duplication results from our design choice to fix equal weights (0.5) on the
    task and constraint clauses, which forces \tlnet to express prioritization between
    the task and constraint by including relevant task predicates in the constraint
    clause.

    Now, consider the Capture and Capture 0 scenarios, which differ in the reward
    to kill the red agent (0.5 vs. 0). While one might naively expect this difference
    to produce different policy explanations, it actually should not change the
    optimal blue agent policies because, to win the game, the blue agent should
    first defend its flag (by killing the red agent), since the red agent uses the
    \rev{\textit{capture}} policy. We see that \tlnet successfully captured this
    nuance by inferring constraints with nearly identical weights,
    reflecting its robustness to minor reward differences that still lead to
    similar policies. NN-TLI and STONE did not infer similar constraints across these
    scenarios. More broadly, this result also demonstrates the importance of inferring
    explanations from actual observed behaviors of RL agents rather than simply looking
    at their specified reward function.
    \begin{table}[tb]
        \centering
        \medskip
        \tabcolsep = 3.6pt
        \caption{\rev{Ablation Study of \tlnet in the CtF Capture Scenario.}}
        \rowcolors{2}{gray!25}{white}
        \begin{tabular}{l|l|l|l|l}
            Features                                                      & Concise.                   & Consist.      & Strict.                    & Accuracy                   \\
            \hline
            Full                                                          & \textbf{0.41}$_{\pm 0.06}$ & \textbf{0.33} & \textbf{0.12}$_{\pm 0.01}$ & \textbf{0.95}$_{\pm 0.02}^*$ \\
            No Filter: $_{S_\mathrm{th}> 1.0}$                            & 0.29$_{\pm 0.07}$          & 0.17          & 0.11$_{\pm 0.01}$          & \textbf{0.95}$_{\pm 0.01}^*$ \\
            No Reg.: $_{\lambda_{R_\mathrm{T}}=\lambda_{R_\mathrm{D}}=0}$ & 0.40$_{\pm 0.09}$          & 0.23          & \textbf{0.12}$_{\pm 0.01}$ & \textbf{0.95}$_{\pm 0.02}^*$ \\
            No Pruning: $_{N_\mathrm{w}=0}$                               & 0.15$_{\pm 0.02}$          & 0.20          & 0.06$_{\pm 0.00}$          & 0.78$_{\pm 0.05}$          \\
            Argmin Negative Traj.                                            & 0.40$_{\pm 0.05}$ & 0.32 & \textbf{0.12}$_{\pm 0.01}$ & \textbf{0.95}$_{\pm 0.01}^*$        \\
        \end{tabular}
        \label{tab:abl_ctf}
    \end{table}


    \subsubsection{Importance of \tlnet Components}
    Finally, the ablation study results shown in \Cref{tab:abl_ctf} show that
    filtering, regularization, and pruning are all important to the performance
    of \tlnet. Removing predicate filtering notably reduced conciseness and consistency, removing regularization decreased conciseness and significantly reduced consistency with no accuracy improvement, and removing pruning largely reduced all metrics.
    \rev{We also tested an argmin approach to sample negative trajectories, where rollouts used the lowest-probability action in the target policy, and saw no significant difference in results compared to random sampling.
    }

    \section{Conclusion}
    We introduce \tlnet, a neuro-symbolic framework for generating wSTL explanations
    of robot policies through predicate filtering, loss regularization, and weight
    pruning. We also propose novel explainability metrics---conciseness, consistency,
    and strictness---to assess explanation quality beyond classification accuracy.
    Experiments in three simulated robotic environments show our method
    outperforms baselines on these metrics while retaining high classification accuracy.
    \rev{Future directions include real-world deployment with online inference (which would require improving sample efficiency and handling predicate uncertainty from sensor noise and partial observability), learning task-constraint weights with a stabilizer, improving negative data generation, exploring post-hoc enforcement of the task-constraint format, and user studies.
    }
    \bibliographystyle{./IEEEtran}
    \bibliography{./IEEEabrv,ref}
\end{document}